\documentclass[english]{cccconf}
\usepackage[comma,numbers,square,sort&compress]{natbib}
\pdfoutput=1
\usepackage{epstopdf}
\usepackage{amsmath, amsfonts}
\usepackage{algorithmic}
\usepackage{algorithm}
\usepackage{array}
\usepackage{booktabs}
\usepackage{multirow}
\usepackage[caption=false, font=footnotesize, labelfont=rm, textfont=rm]{subfig}
\usepackage{textcomp}
\usepackage{rotating}
\usepackage{svg}
\usepackage{pdfpages}
\usepackage{stfloats}
\usepackage{url}
\usepackage{verbatim}
\usepackage{graphicx}
\usepackage{cite}
\usepackage[T1]{fontenc}
\usepackage{bm}
\usepackage{color}
\usepackage{graphicx}
\usepackage{epstopdf}

\usepackage{diagbox}
\usepackage[justification=centering]{caption}

\begin{document}

\title{Time-attenuating Twin Delayed DDPG Reinforcement Learning for Trajectory Tracking Control of Quadrotors}

\author{Boyuan Deng\aref{amss,hit},
        Jian Sun\aref{amss,hit},
        Zhuo Li\aref{amss,hit},
        Gang Wang\aref{amss,hit}}
        
\affiliation[amss]{School of Automation, Beijing Institute of Technology, Beijing 100081, P.~R.~China}
\affiliation[hit]{Key Laboratory of Intelligent Control and Decision of Complex System, Beijing Institute of Technology, Beijing 100081, P.~R.~China
\email{dengboyuan@bit.edu.cn;~sunjian@bit.edu.cn;~zhuoli@bit.edu.cn;~gangwang@bit.edu.cn}}

\maketitle

\begin{abstract}
Continuous trajectory tracking control of quadrotors is complicated when considering noise from the environment. Due to the difficulty in modeling the environmental dynamics, tracking methodologies based on conventional control theory, such as model predictive control, have limitations on tracking accuracy and response time. We propose a Time-attenuating Twin Delayed DDPG, a model-free algorithm that is robust to noise, to better handle the trajectory tracking task. A deep reinforcement learning framework is constructed, where a time decay strategy is designed to avoid trapping into local optima.  The experimental results show that the tracking error is significantly small, and the operation time is one-tenth of that of a traditional algorithm. The OpenAI Mujoco tool is used to verify the proposed algorithm, and the simulation results show that, the proposed method can significantly improve the training efficiency and effectively improve the accuracy and convergence stability. 
\end{abstract}

\keywords{deep reinforcement learning, quadrotor, trajectory tracking}

\footnotetext{The work was supported in part by the National Natural Science Foundation of China under Grants 61925303, 62173034, 62088101.}

\section{Introduction}
Quadrotors have shown great potential for a wide range of applications thanks to their flexibility and maneuverability \citep{ackermann2020ai, loquercio2021learning}. In trajectory-tracking tasks, a control law is to be designed for a quadrotor to track the reference trajectory assigned by the navigation/inertia system. Existing control laws include sliding mode control (SMC) \citep{ma2018flatness}, backstepping control \citep{das2009backstepping}, model predictive control (MPC) \citep{2018PAMPC, 2017Linear, 2016Fast}, decentralized and linear time-variant control \citep{liu2016robust}, and neural network-based control \citep{li2017deep, zhou2017design}. In traditional control algorithms, model-based methods are widely used owing to their good performance \citep{2021Model}. However, these methods heavily depend on accurate dynamical modeling of a quadrotor, and most work only considers quadrotors' dynamical model while ignoring the complex and dynamic environment in practice. Moreover, due to the high-speed movement of quadrotors, any slight disturbance, model mismatch, and a long control interval may lead to a catastrophic crash. As a result, trajectory tracking control of quadrotors is quite challenging in a chaotic environment. Furthermore, a control law needs to be robust against disturbance and adaptive to time-varying reference trajectories in real-time. 

To solve the modeling inaccuracy, researchers have proposed a variety of control methods. For example, M Reinoso et al. \citep{7530409} simplified the four-rotor dynamic model through small angle approximation, thereby reducing the complexity of the SMC design equation. However, there might be chattering at the track reference point. In practical applications, there always exists the so-called balance point jitter, which is fatal to the quadrotor. To solve it, Michael Neuert \citep{2016Fast} has designed a nonlinear MPC framework in milliseconds, which still relies on the model and does not completely solve the problem.

In addition, model-free control methods, such as deep reinforcement learning (DRL), can remedy the problem of excessive dependence on the model.
DRL provides a learning framework to optimize control strategies through interactions with the environment. Thus, it can handle complicated dynamics without the need for modeling \citep{2020Super}. Note that there is always ambient noise during track tracking. To overcome it, Jiying Wu et al \citep{wu2022state} propose a compensation network to the action network of DRL, which can better deal with the noise problem. Nonetheless, the compensation method needs to be further improved to handle complex environmental changes.

\begin{figure}[ht]
	\centering{
	\includegraphics[width=0.9\linewidth,scale=1]{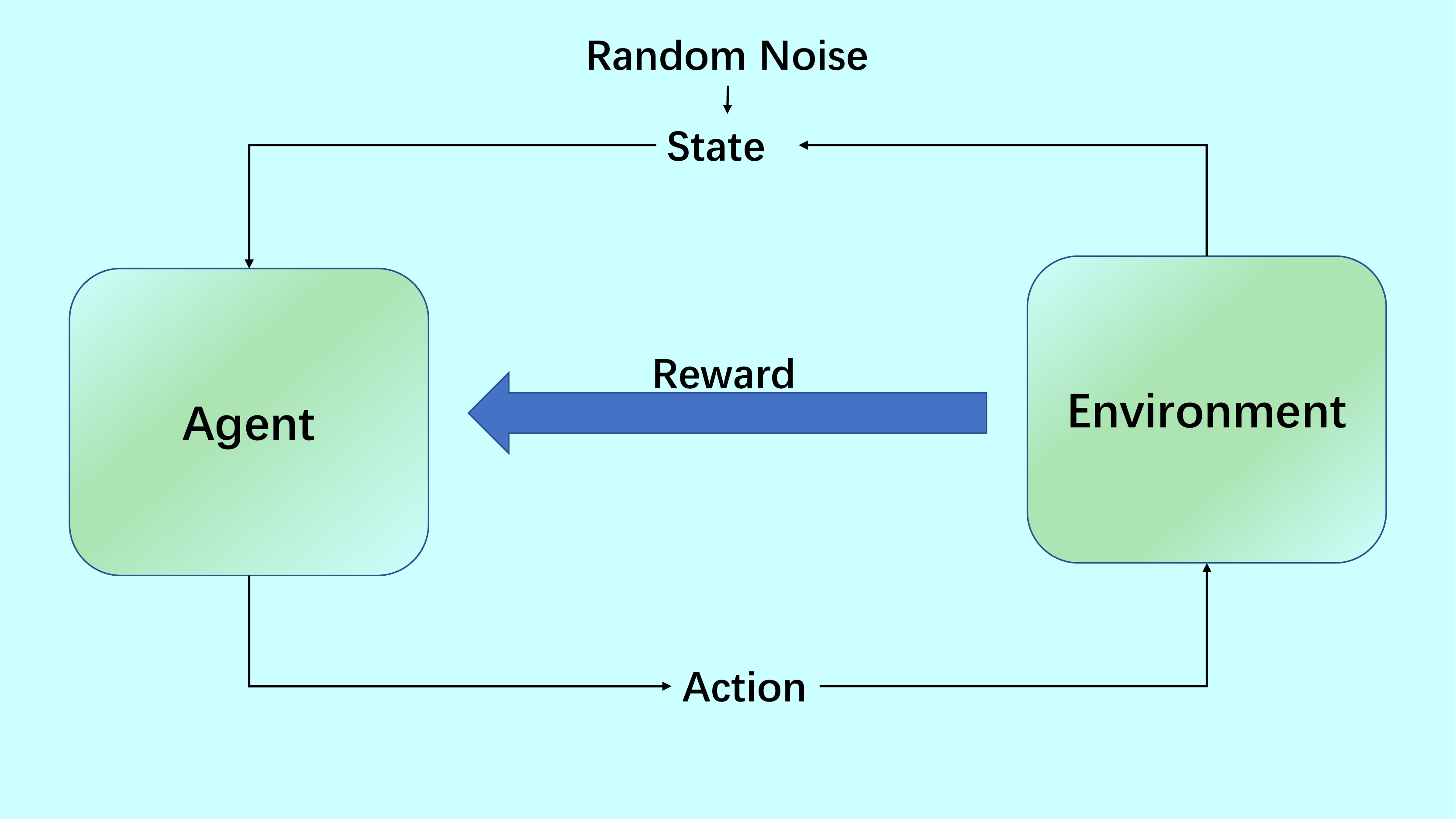}}
    \caption{\centering{Training framework}}
    \label{fig:RL}
\end{figure}

Generally, model-free DRL algorithms such as Deep Deterministic Policy Gradient (DDPG) encourage exploration by injecting action noise, e.g., Gaussian and Ornstein-Uhlenbeck noise, during the training process, optionally following a linear annealing schedule. Different exploration noise lead to different results, or even (potentially poor) local optima. In this paper, by improving the Twin Delayed DDPG (TD3) algorithm, the exploration noise is set to change with the change of time and trend of cycle reward changes, which largely avoids local optimization and makes the algorithm converge quickly and robust to environmental noise, called Time-attenuating Twin Delayed DDPG (T-TD3). Specifically, when designing the simulation environment of the quadrotor trajectory tracking problem, the environment noise network is added (Fig. \ref{fig:RL}), and the T-TD3 algorithm is used to train the neural network controller until convergence. Our research results show that the learning strategy has obtained the tacit knowledge of the risk of near failure during interference, does not rely on the model, and has strong robustness. At the same time, it has made a comparison with the best MPC algorithm in tracking effect and has verified that the T-TD3 algorithm can converge rapidly on the general DRL algorithm verification platform Mujoco. 

\section{Problem Statement}
\subsection{Quadrotor Dynamics}
We model the quadrotor as a rigid body controlled by four motors. The dynamics equations are
\begin{equation}
	\begin{aligned}
		\bm{\dot{p}}_{WB}&=\bm{v}_{WB}                &\bm{\dot{q}}_{WB}&=\frac{1}{2}\bm{\Lambda} (\bm{\omega}_{B}) \cdot \bm{q}_{WB}\\
		\bm{\dot{v}}_{WB}&=\bm{\dot{q}}_{WB}\odot\bm{c}-\bm{g} &\bm{\dot{\omega_{B}}}&=\bm{J^{-1}}(\bm{\eta} -\bm{\omega}_B\times \bm{J}\bm{\omega}_B)\\
	\end{aligned}\label{eq:dynatic}
\end{equation}
where $\bm{p}_{WB}=[p_x, p_y, p_z]^T$ and $\bm{v}_{WB}=[v_x, v_y, v_z]^T$ represent the position and velocity vectors of the quadrotor in the world frame $W$, respectively. We use a unit quaternion $\bm{q}_{WB}=[q_w, q_x, q_y, q_z]^T$ to describe the orientation of the quadrotor and use $\bm{\omega}_B=[\omega_x, \omega_y, \omega_z]^T$ to denote the body rates in the body frame $B$. In addition, $\bm{g}=[0, 0, g_z]$ with $g_z=9. 81m/s^2$ is the gravity vector, $\bm{J}$ is diagonal inertia matrix, $\bm{\eta}$ is the torque, and $\bm{\Lambda} (\bm{\omega}_{B})$ is a centroskew symmetric matrix. Finally, $\bm{c}=[0, 0, c]^T$ is the mass thrust vector. $\bm{p,v,w,q}$ in the following text is the coordinate in the world frame $W$,$\bm{p}(t),\bm{v}(t),\bm{w}(t),\bm{q}(t)$ represents $\bm{p,v,w,q}$ at time t, in the following chapters, we omit subscript for clarity.
\subsection{Trajectory Tracking Problem}
Consider the quadrotor to track a sufficiently smooth trajectory $\bm{p}_d(t):[0, \infty)\to\mathbb{R}^{3}$ with bounded time-derivatives. This work aims to propose a control law steering the quadrotor to pass through a sequence of waypoints along the desired trajectory, which is denoted by $\bm{p}_{di} = \bm{p}_d(iT), i = 1, 2, \cdots, n$ with a time interval $T$. If $T\to 0$ and $n\to \infty$, set $\bm{p}(t)$ as the position of the quadrotor in the world coordinate system at time $t$,then the tracking error $||\bm{p}_d(t)-\bm{p}(t)||$ must converge to a neighborhood of the origin under the proposed control law.

Mathematically, the trajectory tracking problem is summarized as the following optimization problem
\begin{equation}\label{eq:minimize}
	\begin{aligned}
	\mathrm {minimize} &\quad J=\sum_{i=1}^{N}||\bm{p}_i-\bm{p}_{di}||^2+||\bm{u}_i ||^2 , \\
	\mathrm {s. t.} &\quad \omega _{min} \le  \omega_j \le  \omega _{max}, j=x, y, z, \\
	                    &\quad f_{min}\le f_c\le f_{max} , \\
	                    & \quad \text{the dynamics}~ \eqref{eq:dynatic}, 
	\end{aligned}
\end{equation}
where $\bm{p}_i$ denotes the quadrotor's position at the $i$-th time step, $ \bm{u}_i=[f_c, \omega_x, \omega_y, \omega_z]^T$ denotes the vehicle's control input with commanded bodyrates $f_c$ and $\bm{\omega}=[\omega_x, \omega_y, \omega_z]^T$ in \eqref{eq:dynatic}, $\omega _{min}$ and $\omega _{max}$ denote the lower and upper bounds for each element in $\bm{\omega}$, and $f _{min}$ and $f_{max}$ denotes the bounds for $f_c$. 
\section{Methodology}
To solve the trajectory tracking problem in \eqref{eq:minimize}, this section adopts the framework of DRL and proposes a Time-attenuating Twin Delayed DDPG (T-TD3) algorithm. There are two key ingredients in the proposed algorithm: 1) a novel task formulation that combines quadrotor dynamics and smoothness using soft constraints, and 2) a fast-moving convergence strategy to train a policy. 
\subsection{Markov decision process modeling}
Under the framework of RL, the trajectory tracking problem needs to be reformulated as a Markov decision process (MDP). Let a tuple $(\bm{S}, \bm{A}, \bm{R}, \bm{q})$ denote the MDP, where $\bm{S}$ and $\bm{A}$ are observation and action spaces, $\bm{R}$ is the reward space, and $\bm{q}: \bm{S} \times \bm{A} \times \bm{S} \rightarrow \bm{R}$ is the transition probability distribution. 

\textbf{Observation and Action Spaces.} The observation space consists of three main components: the quadrotor’s state at time $t$ $\textbf{s}_{t}^{quad}$ and the information about given paths at time $t$ $\textbf{s}_t^{path}$, and the Euclidean distance error $\textbf{s}_t^{err}$. The quadrotor state is defined as $\textbf{s} _t^{quad}=[\bm{p}_t, \bm{v}_t, \bm{q}_t]\in \mathbb{R}^{10}$, corresponding to the quadrotor’s position, linear velocity and quaternion. To avoid singular expressions in the representation of the rotation, we use a unit quaternion $\bm{q}_t$ to describe the attitude of the quadrotor in each coordinate system.

We define path observation vector as $\bm{s}_t^{path}=[\bm{p}_{d}(t+1), \bm{p}_d(t+1)-\bm{p}(t)] \in \mathbb{R} ^6$, denotes the position of the next moment in a given path and the vector from the current position to it, which contains the spatial direction information. Different from the observation of \citep{attention}, more information will cause more misjudgment in the early training stage.

We define Euclidean distance error scalar as $\bm{s}_t^{error}= ||\bm{p}_d(t)-\bm{p}(t)  ||_2 $, referring to the Euclidean distance between the current position and the position of the next moment in a given path. which renders the agent more sensitive to the output control. 

The action is defined as $\bm{a}_t=[f_c, \omega_x, \omega_y, \omega_z]$, such that the policy can directly map the observation to thrust-angular speed commands. Using direct commands allows quadrotors to have a high-speed response and extreme speed. Moreover, we use normalization in the last layer of policy to control the action within a range. 

\textbf{Reward Function. }Our optimization goal is to minimize tracking error and input. To improve the sensitivity of the agent to tracking errors, we define the error reward $r_e(k)$ as follows
\begin{equation}\label{reward_F}
r_e(k)=\begin{cases}
  ||\bm{p_d}-\bm{p}||_2, &~\text{ if $||\bm{p_d}-\bm{p}||_2^{2}\le 1$, }\\
  ||\bm{p_d}-\bm{p}||_2^{2}, &~\text{ otherwise. } 
\end{cases}
\end{equation}

To ensure a low energy loss, we define the negative input reward $r_u(k)$ as a penalty
\begin{equation}\label{reward_input}
        r_u(k)= - \bm{u}(k)^{T} \Omega \bm{u}(k),
\end{equation}
where the weight matrix $\Omega$ is positive definite. The final reward at each time step $k$ is defined as
\begin{equation}
        r_f(k)=-\rho _1 r_e(k)-\rho _2,
\end{equation}\\
where $\rho _1,\rho _2$ are the weight coefficients.

\subsection{The T-TD3 algorithm}
Generally, random noise is injected into the action $\bm{a}_t$ for the exploration of an optimal policy during the training process. However, in the late training stage, excessive exploration noise causes slow convergence of the algorithm. To resolve this issue, we set the exploration noise to change with time, and the amplitude of change depends on the return of a period. Thus, we propose a time-attenuating version of the TD3 algorithm, called T-TD3 in Algorithm \ref{alg1}.
\begin{algorithm}[ht]
	\caption{T-TD3}
	\begin{algorithmic}[1]
		\STATE {Initialize critic networks $Q_{\theta_1}, Q_{\theta_2}$, and actor network $\pi_\phi $ with random parameters $\theta_1, \theta_2, \phi $ \\}
		\STATE {Initialize target networks $\theta _1{'} \gets \theta_1, \theta_2{'} \gets \theta_2, \phi{'}  \gets \phi$ \\}
		\STATE {Initialize replay buffer $\mathcal{B}$}
		\STATE {\textbf{for} $t= 1$ \textbf{to} $T$ \textbf{do}}
		\STATE \hspace{0.5cm}{ Select action with exploration noise: $a \sim \pi _\phi (s)+\epsilon, $ with $\epsilon \sim \mathcal{N} (0, \sigma )$ and observe reward $r$ and new state $s'$}
		\STATE  \hspace{0.5cm}{Store transition tuple$(s, a, r, s')$ in $\mathcal{B}$}
		\STATE  \hspace{0.5cm}{Sample mini-batch of $N$ transitions$(s, a, r, s')$ from $\mathcal{B} $}
		\STATE  \hspace{0.5cm}{$\tilde{a} \gets \pi _{\phi ^{'}}(s')+\epsilon   , \epsilon \sim clip(\mathcal{N}(0, \sigma ), -c, c)$}
		\STATE  \hspace{0.5cm}{$y\gets r+\gamma min_{i=1, 2}Q_{\theta _{i}{'}}(s{'}, \tilde{a} ) $}
		\STATE  \hspace{0.5cm}{Update critics $\theta _i\gets argmin_{\theta _i}N^{-1}\sum (y-Q_{\theta _i}(s, a))^2$}
		\STATE  \hspace{0.5cm}{Update explore-range}
		\STATE  \hspace{0.5cm}{$\sigma \gets \beta u_{max}e^{-\lambda t}+(1-\beta )u_{max}e^{-\delta  }$}
		\STATE  \hspace{0.5cm}{$\delta=\sum_{i\in mini-batch}\frac{r_i-r_{min}}{r_{max}-r_{min}}$}
		\STATE  \hspace{0.5cm}{\textbf{if} $t$ mod $d$ \textbf{then}}
		\STATE  \hspace{1cm}{Update $\phi$ by the deterministic policy gradient:}
		\STATE  \hspace{1cm}{$\bigtriangledown _\phi J(\phi)=N^{-1}\sum \bigtriangledown_aQ_{\theta _1}|_{a=\pi _{\phi }(s)}\bigtriangledown _{\phi }\pi _\phi (s)$}
		\STATE  \hspace{1cm}{Update target networks:}
		\STATE  \hspace{1cm}{$\theta _{i}{'}\gets \tau \theta _i+(1-\tau )\theta _{i}^{'}$}
		\STATE  \hspace{1cm}{$\phi^{'}\gets \tau \phi+(1-\tau )\phi^{'}$}
		\STATE  \hspace{0.5cm}{\textbf{end if}}
		\STATE  {\textbf{end for}}
	\end{algorithmic}
	\label{alg1}
\end{algorithm}

The training phase starts in line 4. The agent interacts with the environment and stores training data in lines 5-7, $a$ represents action,$\epsilon$ represents the added exploration noise, $\mathcal{N} (0, \sigma )$ represents Gaussian noise with mean value of 0 and variance of $\sigma$. The action to obtain the corresponding status according to the next status in line 8. The discount return is obtained in line 9, and the critic network is updated in line 10.

Our algorithm enforces the notion that similar actions should have similar values. Thus, we modify the update of exploration noise in lines 11-13, where the first part decays exponentially over time, and the second part increases/decreases the extent of exploration according to the return over a period of time to prevent action values from being underestimated, and keep the target close to the original action, $\beta,\lambda$ are constants. The network parameters are updated in lines 15-19.

\textbf{Policy Architecture} The neural network framework uses an end-to-end policy, the observation space to action space corresponds to a given path to quadrotor input. 
\begin{figure}[ht]
	\centering
	\includegraphics[width=0.9\linewidth]{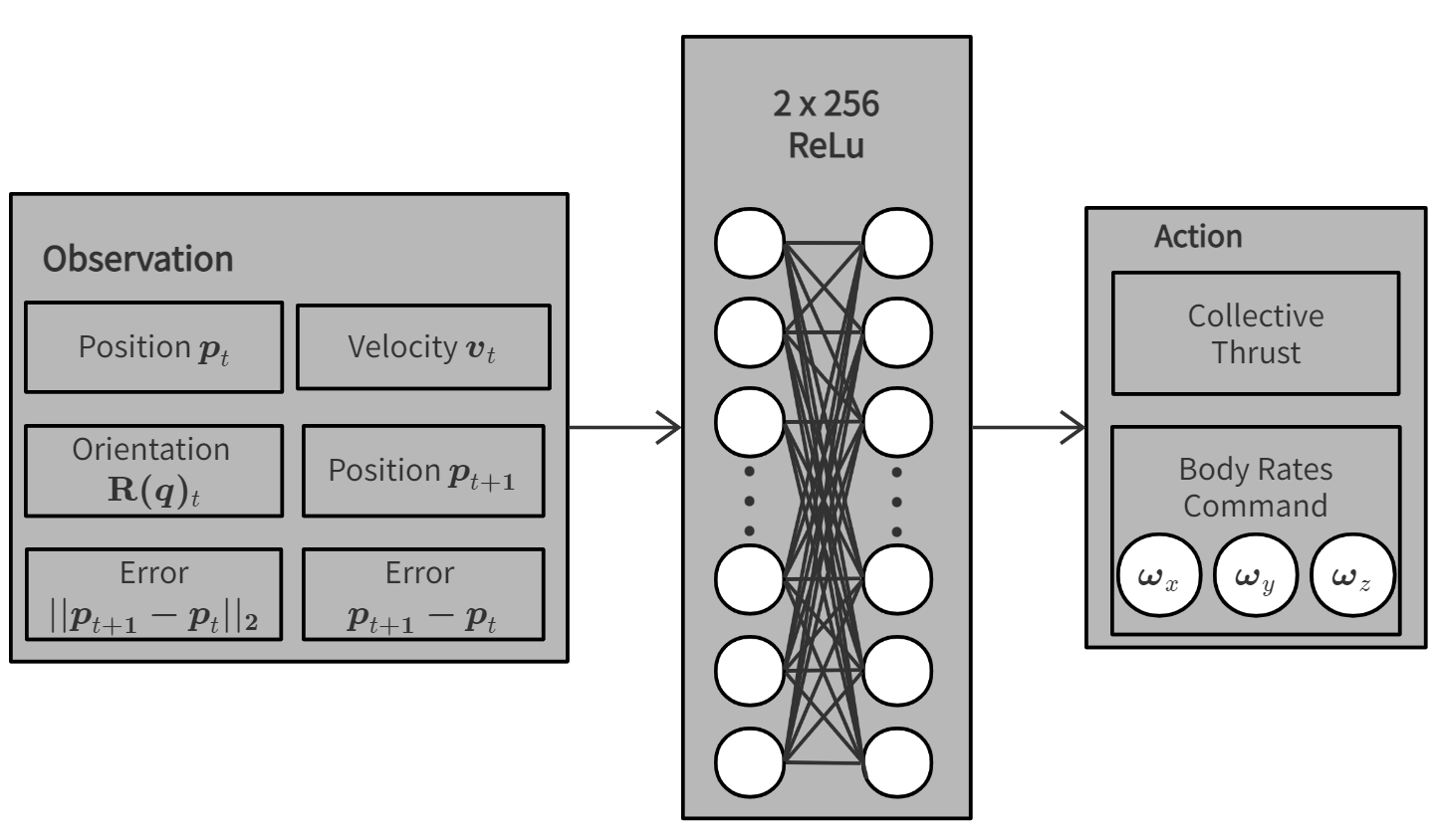}
	\caption{Illustration of the policy architecture, including observation and action. }\label{fig:architecture}
\end{figure}
Fig \ref{fig:architecture} illustrates our architecture, including observation and action spaces, which is a 2-layer multilayer perception (MLP).

\textbf{Training Strategy} TD3 algorithm has good benchmark performance in the continuous control task, because the tracking task has high precision and robustness. Nonetheless, our task for the TD3 algorithm is challenging due to the slow convergence at the late stage of the training process. In contrast, the training process of our T-TD3 algorithm has two key points, one is to speed up the training process by reducing the noise index over time, and the other is to evaluate the cycle reward to prevent overfitting, which enables us to achieve perfect tracking performance under any trajectory. 

\begin{figure}[ht]
	\centering{
	\includegraphics[width=0.8\linewidth]{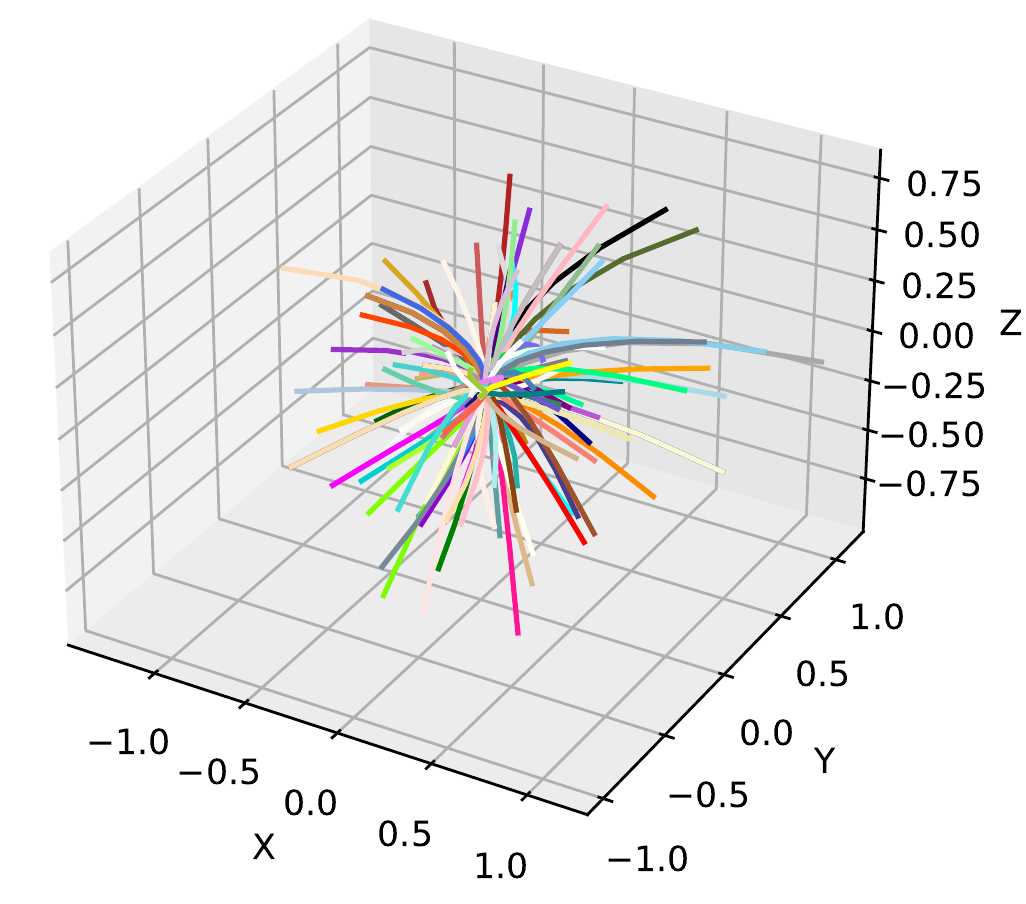}}
    \caption{\centering{Random initial diversification data during the training process. }}\label{fig:path}
\end{figure}

To cope with possible complex trajectories in practical environments, we use the dynamics in the initial simulation to randomly generate diversified data, and use a 4th-order Runge-Kutta scheme for the numerical integration of the dynamic equation in Fig \ref{fig:path}. Moreover, we use 20 parallel agent-sharing policies, collect the data in simulation, and update the policy online, which enriches the empirical state and observations of the agent, and results in a significant speed-up of the data collection process.
All policies are trained on the server with Intel Xeon Gold 5218R CPU and $4\times$GeForce RTX 3090.
\section{Experimental results}
Our proposed algorithm answers experiment research questions: (i) How effective is tracking in non-training trajectory? (ii) How is the energy loss compared to other algorithms? (iii) What is the running time of our learning-based strategy? (iv) How does our algorithm compare with other algorithms in the MUJOCO environment?

In order to verify the effectiveness of the algorithm, we designed two trajectories for testing. The 4.1 part interprets how the tracks are generated, and the 4.2 part interprets the above four issues.

The physical performance of the quadrotor is summarized in Table\ref{tab:qua_parama}, and the hyperparameters of the proposed T-TD3 algorithm are summarized in Table \ref{tab:alg_parama}.
\begin{table}[!htb]
  \centering
  \caption{Parameters of the quadrotor dynamic equation}
  \label{tab1}
  \begin{tabular}{l|l}
    \hhline
    $f_{max}[N]$          & 20 \\ \hline
    $f_{min}[N]$   & 0 \\ \hline
    $\omega_{max}[rad/s]$   & 6 \\ \hline
    $\omega_{min}[rad/s]$         & -6 \\ \hline
    $m[kg]$         & 1.5 \\ \hline
    \hhline
  \end{tabular}
   \label{tab:qua_parama}
\end{table}
\begin{table}[!htb]
  \centering
  \caption{Parameters of the T-TD3}
  \label{tab1}
  \begin{tabular}{l|l}
    \hhline
    learning rate          & 0.9995 \\ \hline
    batch size   & 256 \\ \hline
    $\gamma$   & 0.99 \\ \hline
    $\lambda$         & 0.01 \\ \hline
    $\beta $         & 0.08 \\ \hline
    d         & 2 \\ \hline
    c         & 0.1 \\ \hline
    $u_{max}$         & 1 \\ \hline
    \hhline
  \end{tabular}
   \label{tab:alg_parama}
\end{table}
\subsection{Design for trajectory}
To demonstrate the generality of our algorithm, two different reference trajectories are adopted as follows:

1) Spiral-RT:
\begin{equation}
    \begin{aligned}
        &x_d(t) = 0.1t \text{cos}( \frac{\pi}{20}t),\\
        &y_d(t) = 0.1t \text{sin}( \frac{\pi}{20}t),\\
        &z_d(t) = 0.1t.
    \end{aligned}\label{SpiralRT_equ}
\end{equation}
2) The LOS-based generated reference trajectory (LOS-RT) :
\begin{equation}
    \begin{aligned}
        &x_d(t+1)=x_d(t)+cos(t)p(t),\\
        &y_d(t+1)=y_d(t)+sin(\theta (t))p(t),\\
        &z_d(t+1)=z_d(t)+0.0001,
    \end{aligned}\label{LOSRT_equ}
\end{equation}
where $\theta(t)$ and $p(t)$ are sampled from uniform distribution $U(\cdot , \cdot )$ every $T$ time steps, as follows: 
\begin{equation}
    \begin{aligned}
        &\theta (t)=\left\{
    \begin{aligned}
    &U(\theta _{min}, \theta _{max} ), \quad &  \frac{t}{T}, k=0, 1. . . , N-1. \\
    &\theta (kT), \quad & t\in (KT, (k+1)T), \\
    \end{aligned}
    \right. \\
    &p(t)=\left\{
    \begin{aligned}
    &U(d_{min}, d_{max}), \quad  &\frac{t}{T}, k=0, 1. . . , N-1, \\
    &p(kT), \quad &t\in (KT, (k+1)T). \\
    \end{aligned}
    \right. 
    \end{aligned}
\end{equation}
The initial two positions of LOS-RT are set as $\bm{p_d}(0) = (0, 0, 0)$ and $\bm{p_d}(1) = (0.1, 0.1, 0.1)$. The next reference point $\bm{p}_d(t+1) = (x_d(t+1), y_d(t+1), z_d(t+1))$ is generated based on the current reference point $\bm{p_}d(t) = (x_d(t), y_d(t), z_d(t))$, expected deflection angle $\theta(t)$ and expected sailing distance $\bm{p}(t)$, which is inspired by LOS. Set as $\theta_{min}=-(\pi/120)$, $\theta_{max}=(\pi/120)$, $p_{min} = 1. 5$, and $p_{max} = 2. 5$. In this situation, the designed controller needs to address multiple kinds of horizontal reference trajectories. \par
The step length of the reference trajectory is $T_{length} = 3$ and sample time is $T = 0.001$s, which means that the reference trajectory is tracked in 3s.
\subsection{Indicator interpretation}
\emph{A. Tracking error}

\begin{figure}[ht]
	\centering
        \includegraphics[width=0.8\linewidth]{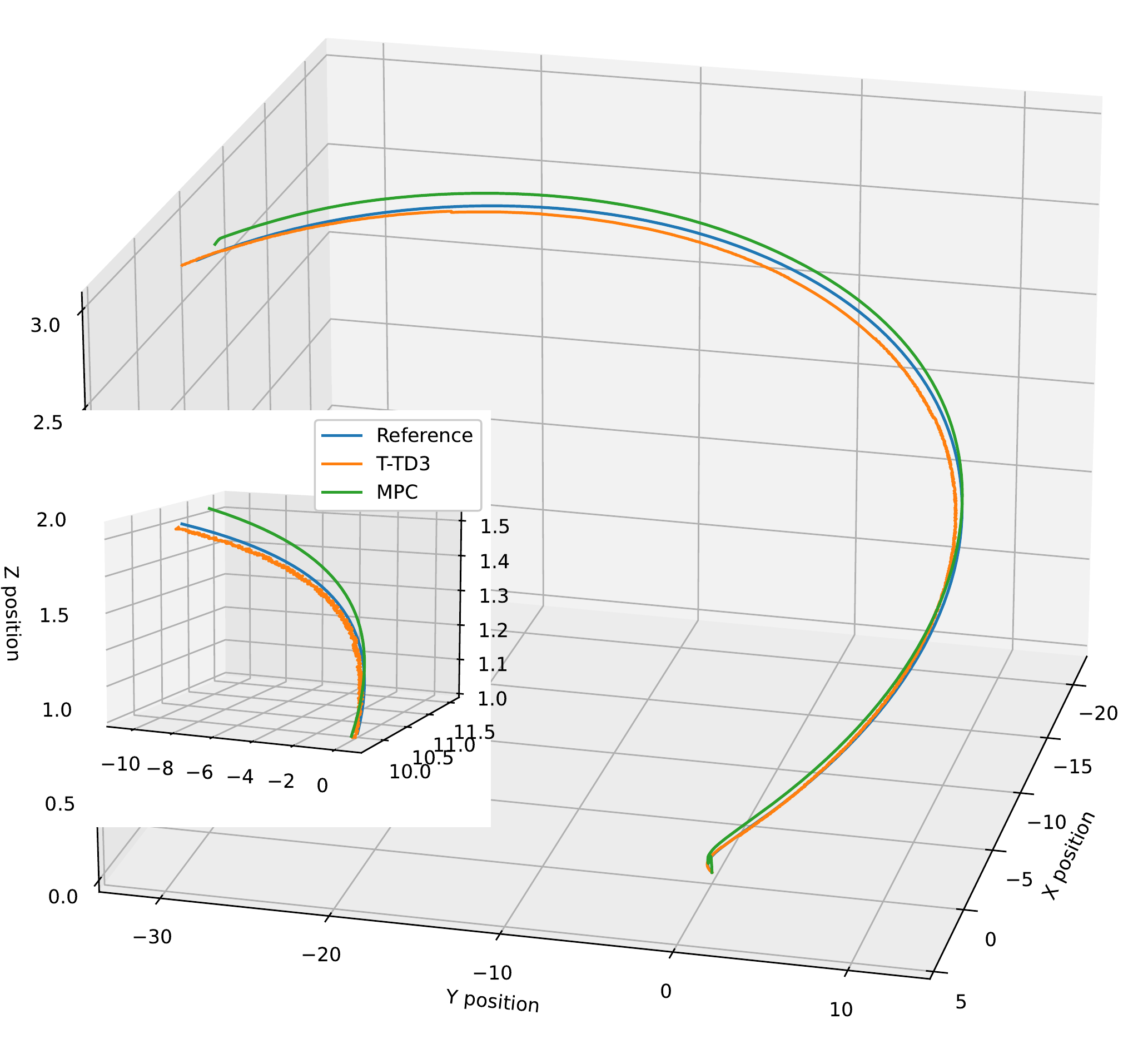}
	\caption{Given SpiralRT trajectory, the MPC method is compared with our method.}\label{Fig:SpiralRT}
\end{figure}

\begin{figure}[ht]
	\centering
        \includegraphics[width=0.8\linewidth]{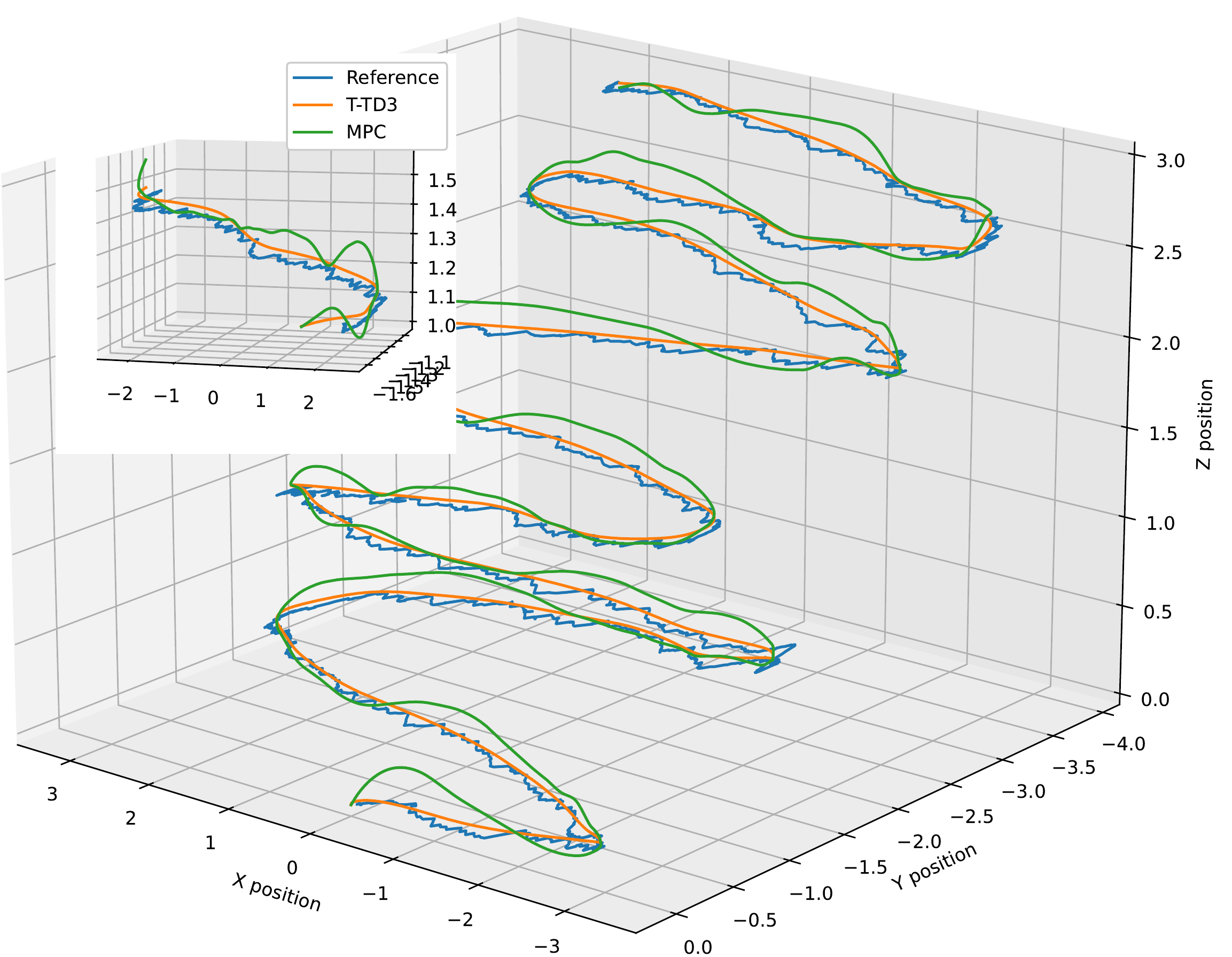}
	\caption{Given LOSRT trajectory, the MPC method is compared with our method.}\label{Fig:LOSRT}
\end{figure}

We compared the trained MLP controller with the MPC on two curves. The running tracks of the two algorithms with the reference tracks are given in the Fig\ref{Fig:SpiralRT}, Fig\ref{Fig:LOSRT}. The calculated tracking error is shown in the Table \ref{Tabel:Track_error}, on the SpiralRT trajectory, the error of our method is only 33.12\% of that of MPC, on the LOSRT trajectory, the error of our method is 73.44\% of that of MPC, the data shows that our method is better in tracking effect.

\begin{table}[!htb]
  \centering
  \caption{Comparison of Tracking Error}
  \begin{tabular}{c|c|c}
    \hhline
    \diagbox{Trajectory}{Method}           & T-TD3  & MPC\\ \hline
    SpiralRT         & 25.99 & 78.47 \\ \hline
    LOSRT       & 398.74 & 542.92 \\ \hline
    \hhline
  \end{tabular}
  \label{Tabel:Track_error}%
\end{table}
\emph{B. Energy Loss}
\begin{figure}[ht]
	\centering
	\subfloat[SpiralRT]{\includegraphics[width=0.45\linewidth]{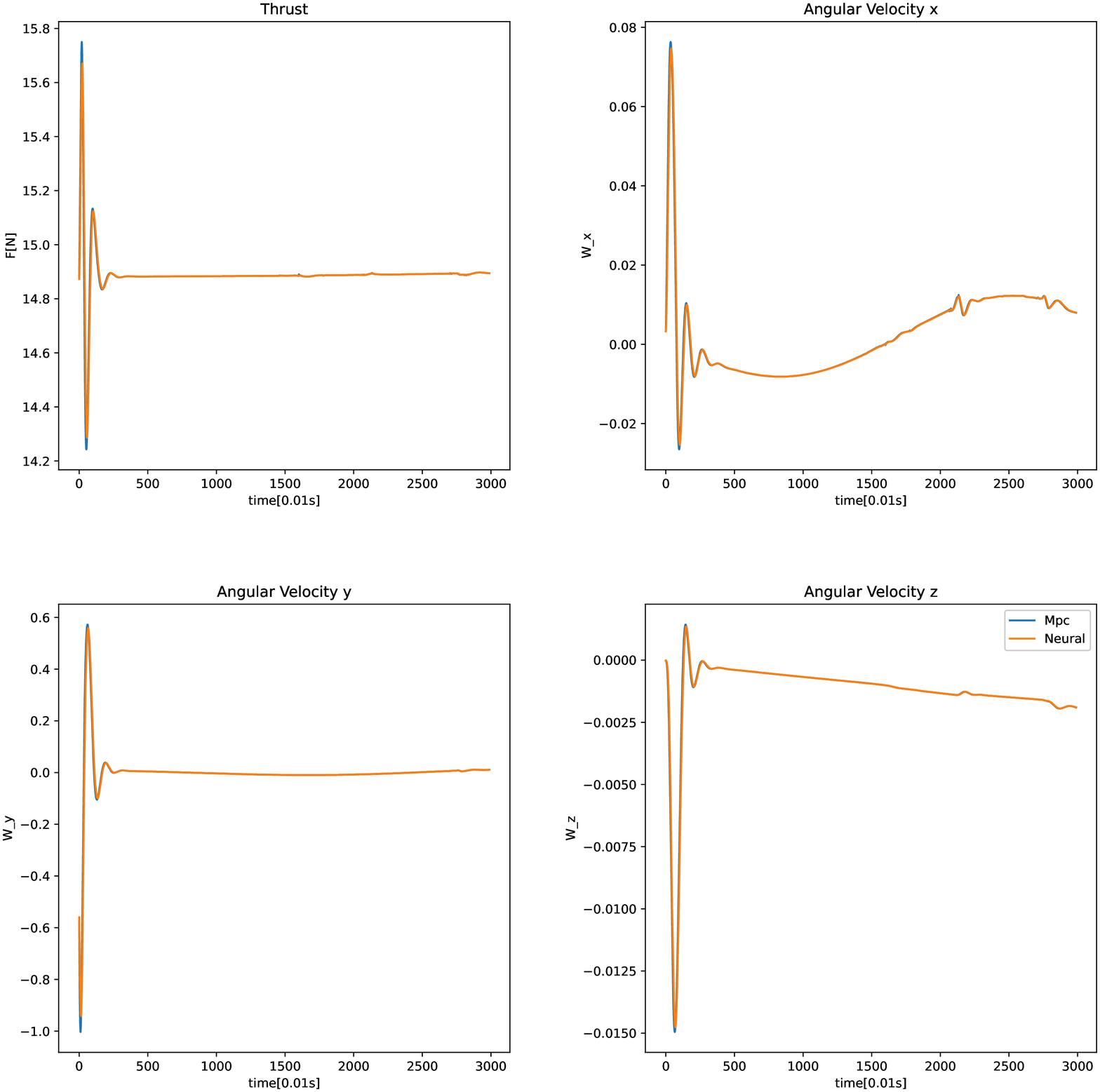}}\hspace{5pt}
	\subfloat[LOSRT]{\includegraphics[width=0.45\linewidth]{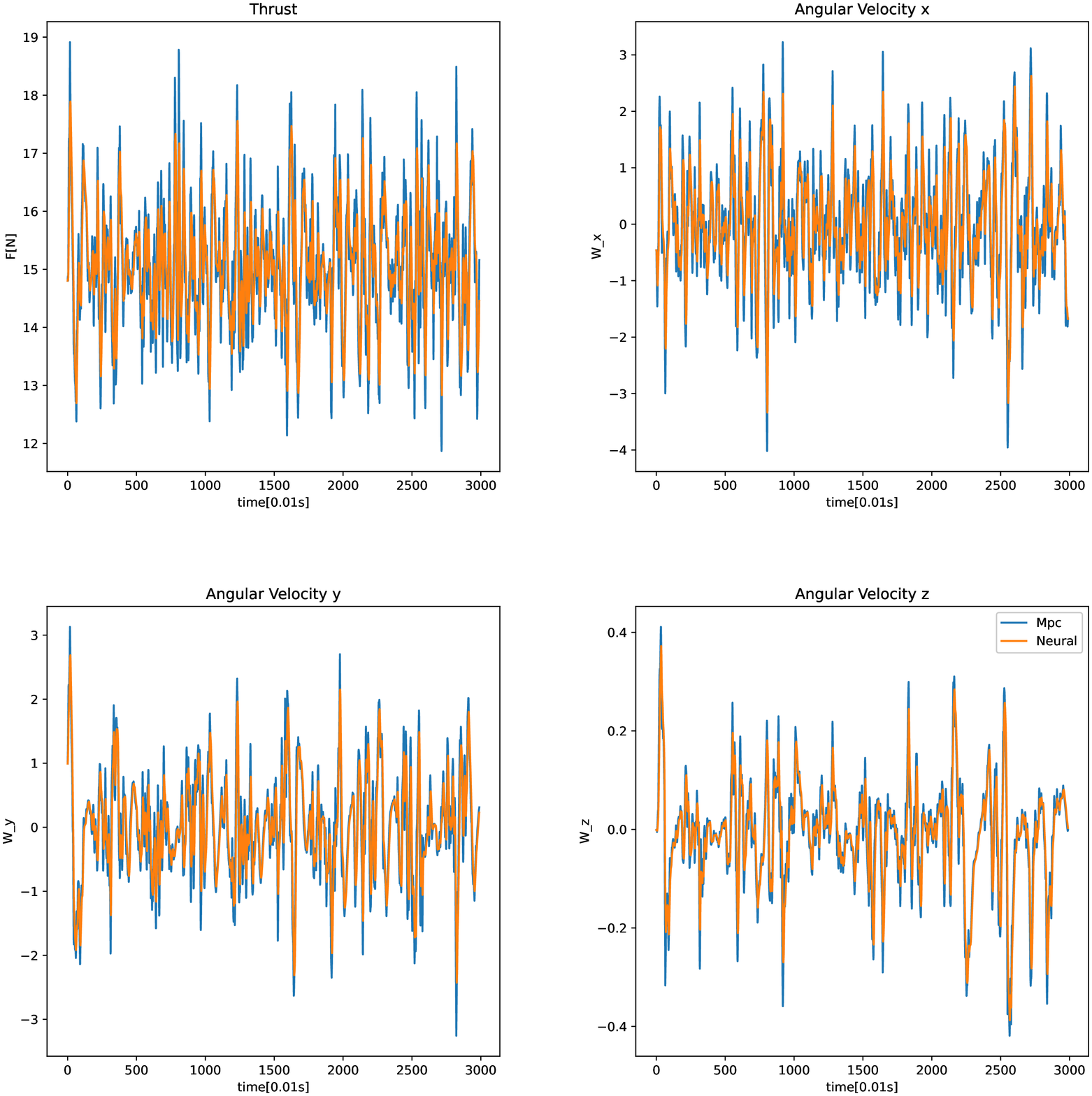}}\\
	\caption{\centering{Energy loss of MPC method and Neural method under two test tracks respectively}}\label{Fig:Energy_Loss}

\end{figure}

Under the two random trajectories tested, we recorded the change curve of u, whether it is our strategy or MPC method, the input of u is normalized, and the value range in $[-1, 1]$. Generally, the loss is defined as 
$$
Energy=u^Tu.
$$

To simplify the calculation and make the data comparison clearer, we used the following loss function
$$
Energy=\sqrt{u^Tu}.
$$

When the error between the trajectory and the meta trajectory obtained by the MLP controller is small, the output is smaller, which means that the MLP controller is more excellent. See Fig\ref{Fig:Energy_Loss} and Tabel \ref{Tabel:Energy_loss}, the data shows that our method has lower energy loss under the trajectory with high complexity.

\begin{table}[!htb]
  \centering
  \caption{Comparison of Energy Loss}
  \begin{tabular}{c|c|c}
    \hhline
    \diagbox{Trajectory}{Method}           & T-TD3  & MPC\\ \hline
    SpiralRT         & 2759. 582772 & 2762. 378835 \\ \hline
    LOSRT       & 7156. 452883 & 7467. 541158 \\ \hline
    \hhline
  \end{tabular}
  \label{Tabel:Energy_loss}%
\end{table}
\emph{C . Running Time}

To avoid the accident of the experiment, we recorded the time required for the MPC controller and T-TD3 method to run three times respectively in the whole time cycle. It should be noted that the control cycle during the test is 0.001 second, and the single time length of a track is 3 second. Therefore, the controller has a total output of 3000 times.The time we recorded is the average of the three Monte-Carlo experiments. We put the time into the Table \ref{Running Time}, the data shows that our method reduces the operation time, which is 4 times less than MPC.
\begin{table}[h]
  \centering
  \caption{The running time(s) of the MPC controller is compared with that of the T-TD3 controller.}
  \begin{tabular}{c|c|c}
    \hhline
    \diagbox{Trajectory}{Method}           & T-TD3  & MPC\\ \hline
    SpiralRT         & 4.879404 & 19.360494 \\ \hline
    LOSRT       & 4.754059 & 16.305317 \\ \hline
    \hhline
  \end{tabular}
  \label{Running Time}%
\end{table}

\emph{D. Comparison With TD3 Algorithm}

To verify the effectiveness of the proposed algorithm, we ran our algorithm on the general platform MUJOCO for reinforcement learning algorithm verification. In order to avoid the contingency of the experiment, we conducted three experiments, each with different random seeds. The results are shown in Fig \ref{al_result}:

\begin{figure}[h]
	\centering
	\subfloat[Ant-v2]{\includegraphics[width=0.45\linewidth]{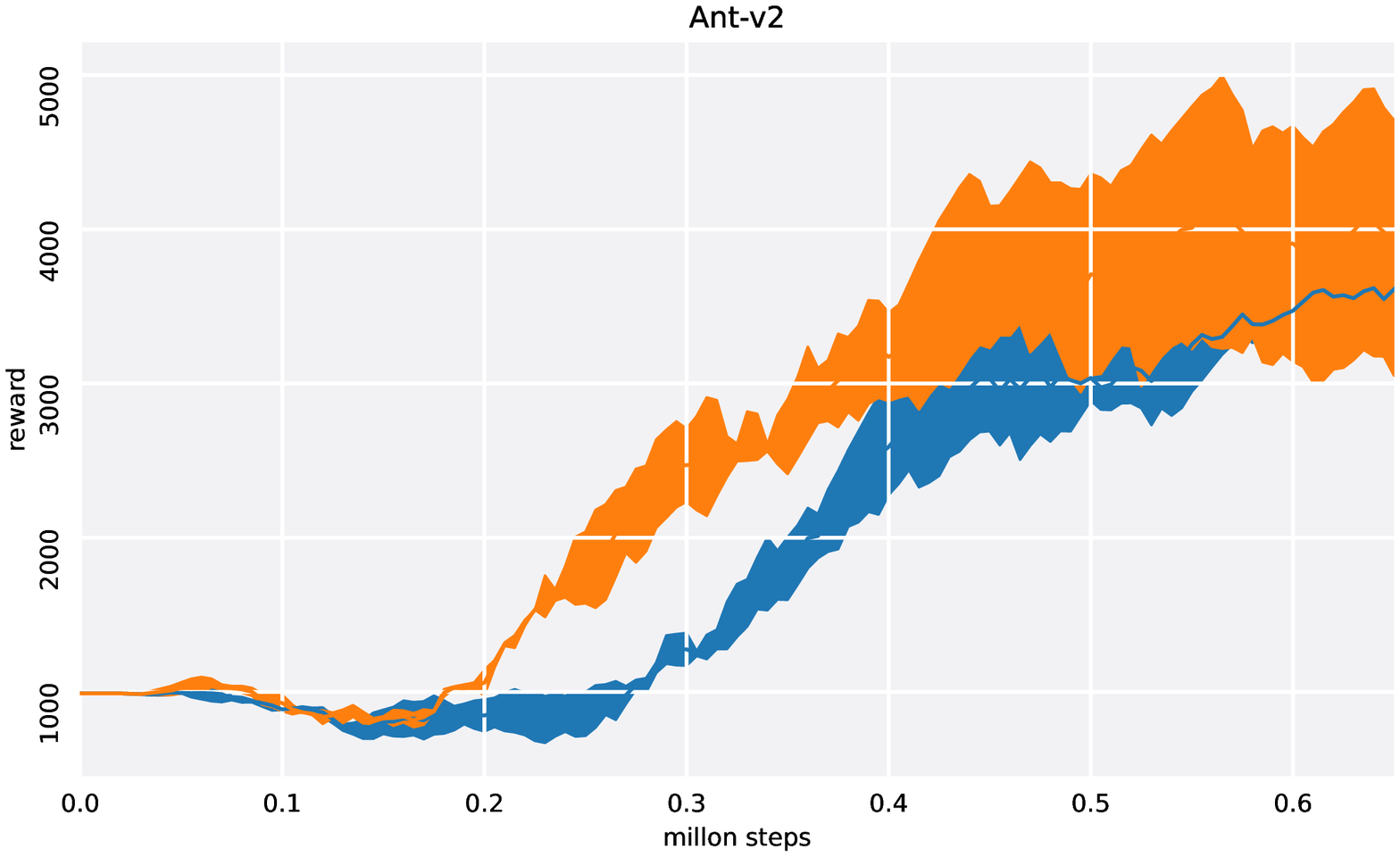}}\hspace{5pt}
	\subfloat[Walker2d-v2]{\includegraphics[width=0.45\linewidth]{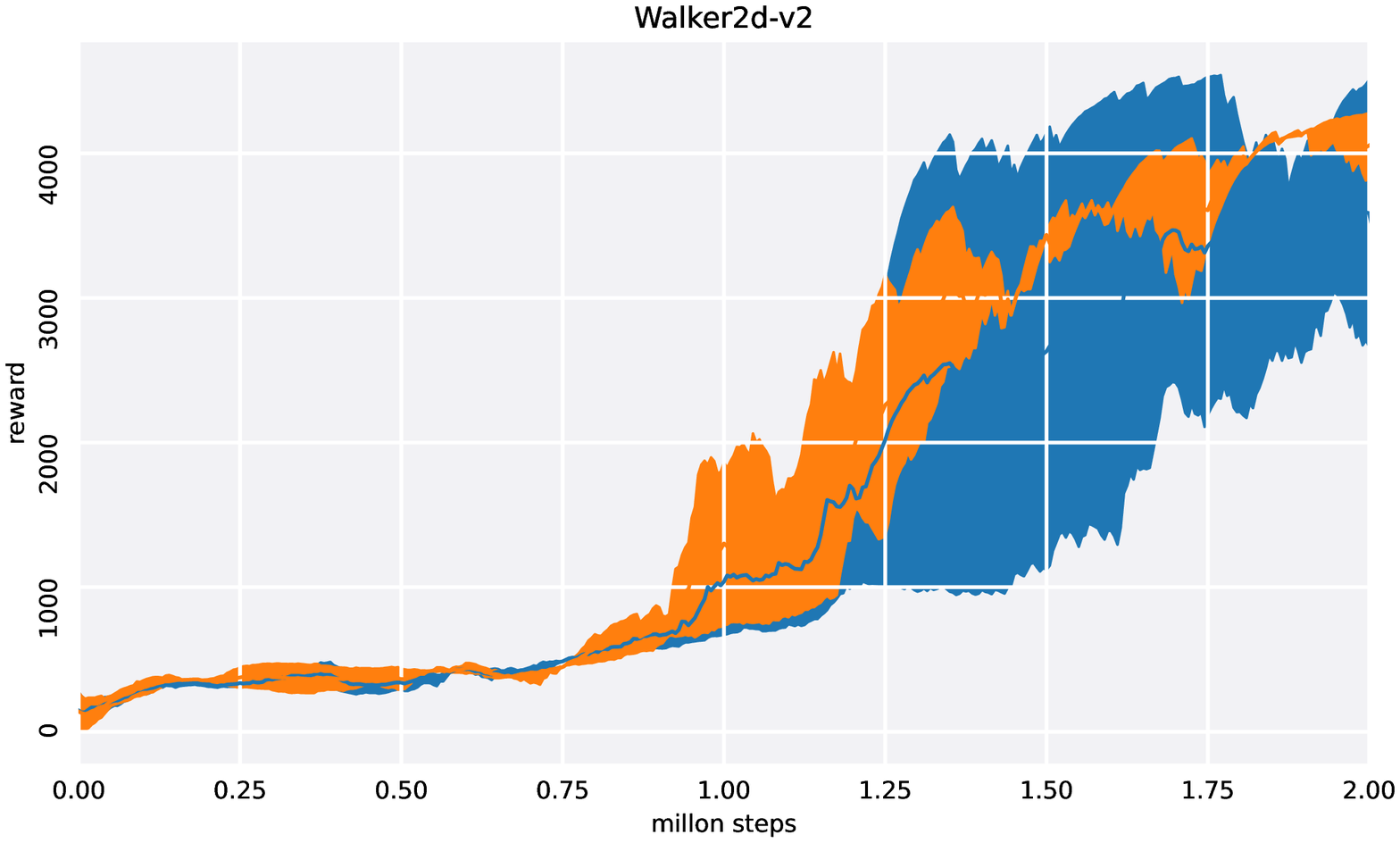}}\\
	\subfloat[InvertedDoublePendulum-v2]{\includegraphics[width=0.45\linewidth]{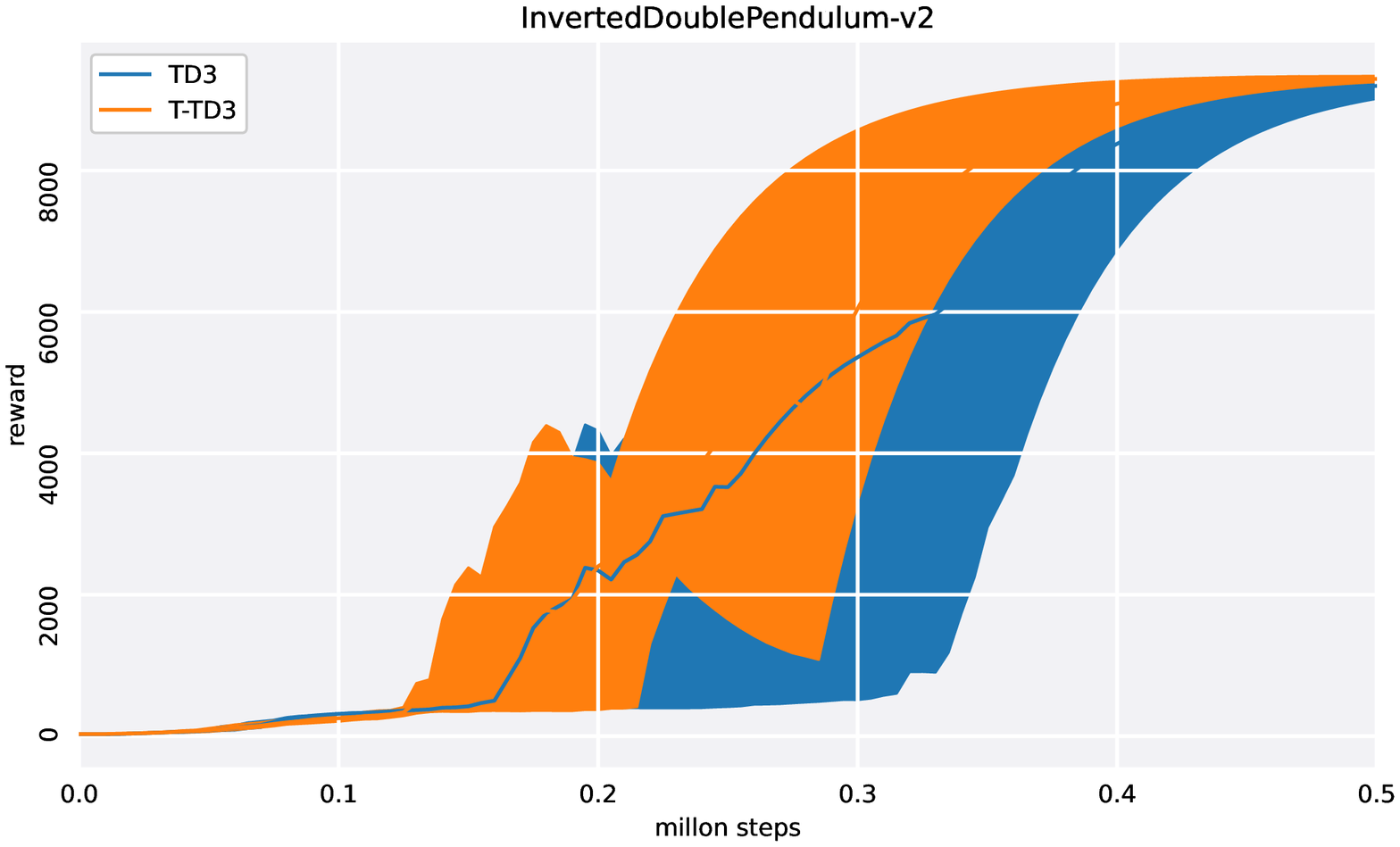}}\hspace{5pt}
	\subfloat[Hopper-v2]{\includegraphics[width=0.45\linewidth]{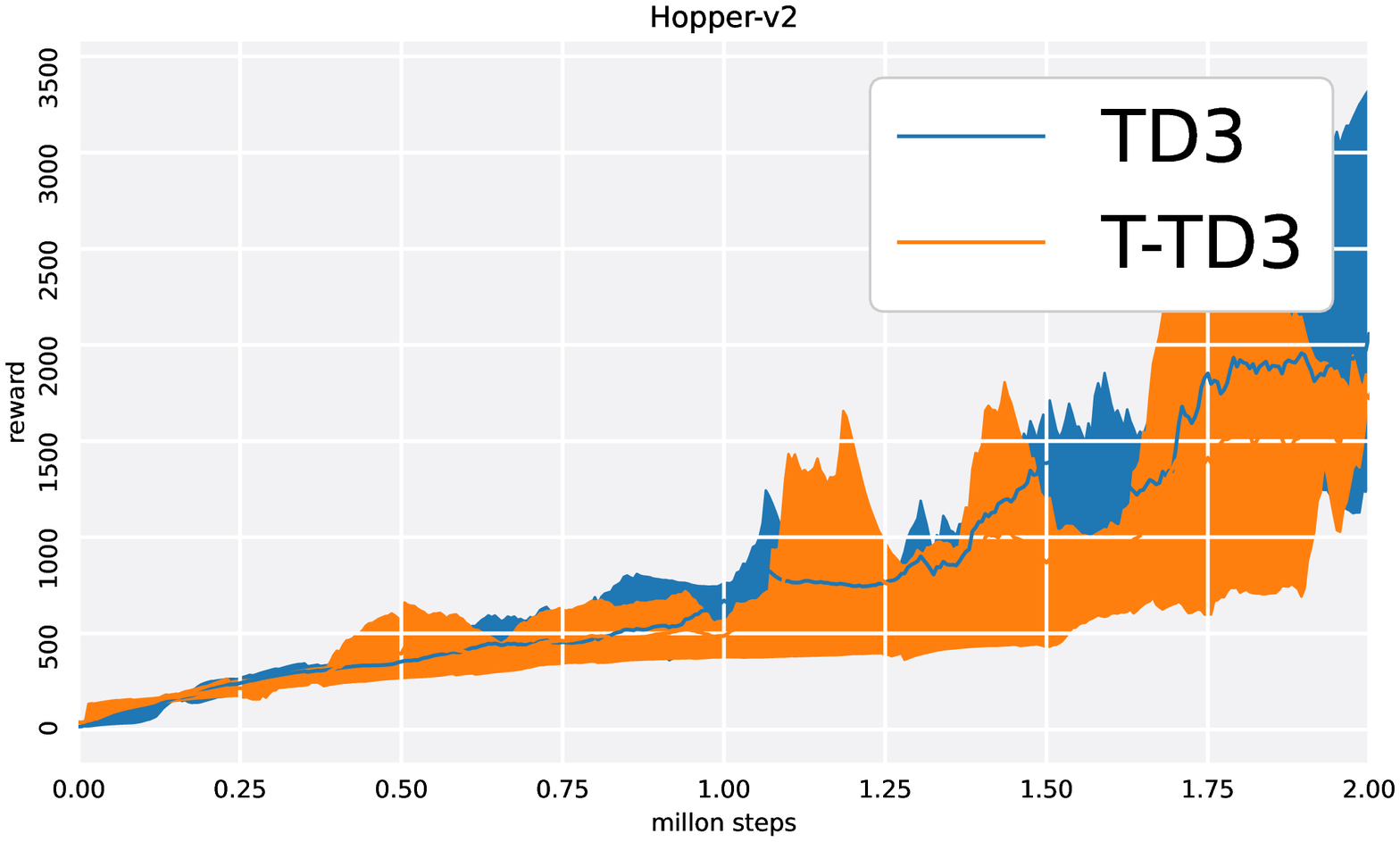}}
	\caption{Learning curves for the OpenAI gym continuous control tasks. The shaded region represents half a standard deviation of the
		average evaluation over 3 random trials. Curves are smoothed uniformly for visual clarity.}
  \label{al_result}
\end{figure}

Through three experiments in four environments and recording single-step returns, the data shows that our method converges faster in Ant-v2, Walker2d-v2 and InvertedDoublePendulum-v2 environments, and is slightly worse than TD algorithm in 4 environments.
\section{Conlusions}
In this paper, a reinforcement learning method based on the TD3 algorithm is proposed to train MLP controller to realize quadrotor control under the trajectory tracking problem. We used MLP to solve the high-dimensional continuous control problem, and generated strong disturbance rejection, high agility and strong robustness controllers through random trajectory pre-training. More importantly, we proved that the trained MLP controller can adapt online and achieve a good tracking effect for different trajectories. Compared with MPC, our method reduces tracking error, and greatly reduces the operation time, which is 4 times less than MPC.

\bibliographystyle{ieeetr}
\bibliography{cite/cite.bib}
\end{document}